\documentclass{article} %
\usepackage{preprint,times}

\usepackage{amsmath,amsfonts,bm}

\def\eqref#1{equation~\ref{#1}}

\def\1{\bm{1}}

\DeclareMathAlphabet{\mathsfit}{\encodingdefault}{\sfdefault}{m}{sl}
\SetMathAlphabet{\mathsfit}{bold}{\encodingdefault}{\sfdefault}{bx}{n}

\usepackage{hyperref}
\usepackage{url}
\usepackage{graphicx}
\usepackage{booktabs}
\usepackage[table]{xcolor}
\definecolor{oursblue}{RGB}{225,245,248}
\usepackage{amssymb}
\usepackage{pifont}

\usepackage{float}
\usepackage{wrapfig}

\title{BridgeAfford: Bridging Query–Target Gap for Generalizable 3D Affordance Grounding
}

\author{Yijie Su\textsuperscript{1}, \ 
Yakun Zhu\textsuperscript{1}, \ 
Yi Bin\textsuperscript{1,*}, \ 
Pengpeng Zeng\textsuperscript{1}, \ 
Liang Chen\textsuperscript{1}, \ 
Guoqing Wang\textsuperscript{2}, \\
\textbf{Jingkuan Song\textsuperscript{1},} \ 
\textbf{Heng Tao Shen\textsuperscript{1}}
 \\[4pt]
\textsuperscript{1}Tongji University \quad
\textsuperscript{2}University of Electronic Science and Technology of China \\
\textsuperscript{*}Corresponding author: Yi Bin
(\texttt{yi.bin@hotmail.com})
}

\preprintfinalcopy %
\begin{document}

\maketitle
\fancyhead{}                      %

\begin{abstract}

Language-guided 3D affordance grounding aims to localize functional regions on 3D objects according to interaction intents. However, existing methods exhibit poor generalization to unseen objects. One challenge is the query–target gap: a semantic discrepancy between query features encoding interaction intents and target-region features, leading to inaccurate grounding. 
End-to-end training favors query representations tailored to seen objects, limiting their compatibility with shifted target-region features on unseen objects.
To address this issue, we propose BridgeAfford, a framework that reduces the query–target semantic gap from two complementary perspectives. First, Interaction Region Alignment (IRA) promotes cross-object alignment among geometrically similar interaction regions during training. This enables structures on unseen objects that resemble previously observed ones to produce representations similar to those learned during training, facilitating target-region recognition. Second, Visual Evidence Aggregation (VEA) leverages a 2D affordance model to refine the attention distribution and guide the query toward the correct region, yielding more target-aware queries. Extensive experiments on the LASO and PIAD datasets demonstrate that BridgeAfford outperforms existing methods, particularly on unseen objects, highlighting its robust generalization ability.
\end{abstract}

\section{Introduction}

3D affordance grounding aims to localize functional regions on 3D objects according to interaction intents, which is fundamental for embodied agents. Recently, language-guided 3D affordance grounding has attracted increasing attention, as natural-language instructions can express more flexible and fine-grained interaction intents than predefined affordance labels~\citep{laso}. 

However, existing methods still suffer from limited generalization, with substantial performance degradation on objects unseen during training. Most prior approaches~\citep{laso,geal,glance} follow a common LASO paradigm: given an instruction and a point cloud, the model uses the interaction intent to extract target-region information from point-wise object features and incorporate it into query features, aligning them semantically with the target region. The affordance mask is subsequently predicted by measuring the similarity between the target-aware queries and the point-wise object features.

For objects seen during training, the model can learn and memorize the correspondence between interaction intents and target regions, incorporating the correct regional features into the query to achieve semantic alignment with the target region. On unseen objects at inference time, however, it may encounter interaction-region representations that deviate from those observed during training. As a result, it cannot reliably use learned associations to incorporate the relevant information into the query, leading to a semantic gap with the target region.

\begin{figure}[t]
\centering
\includegraphics[width=\linewidth]{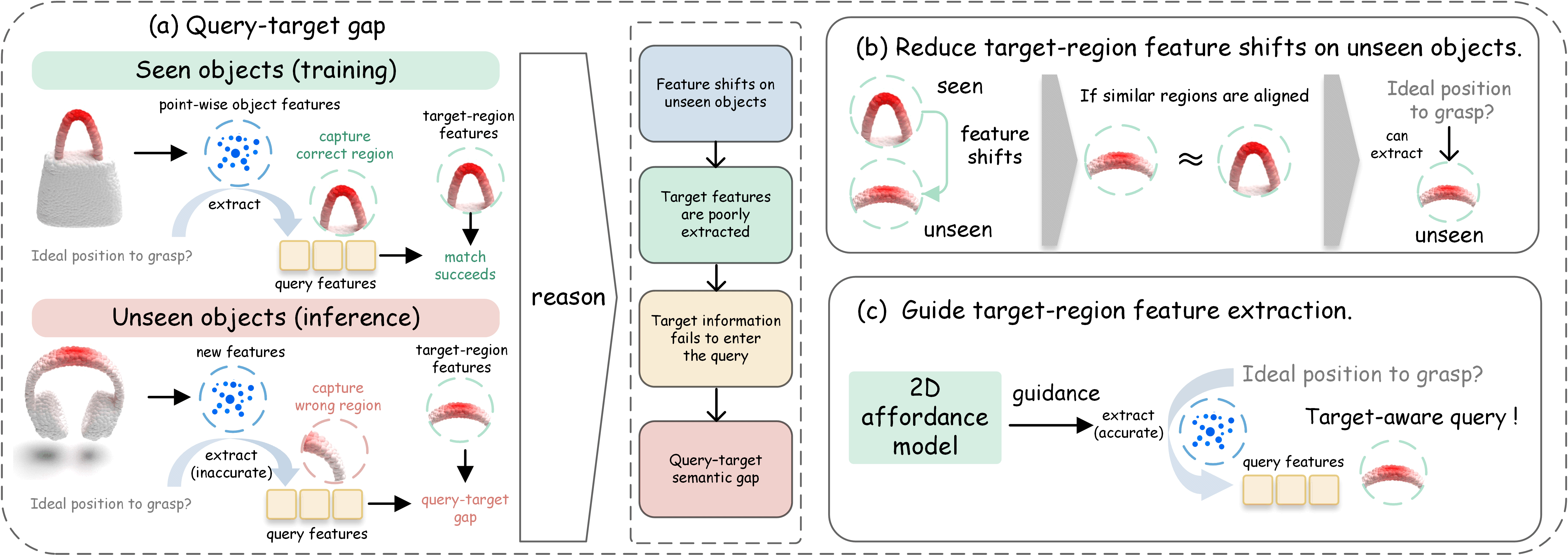}
\vspace{-2mm}
\caption{
Motivation for BridgeAfford.  
(a) Accurate prediction relies on high semantic similarity between the query and the target-region representation. However, shifted local representations on unseen objects may prevent the model from extracting the target region, resulting in query features with low similarity to it.
(b) Unseen objects may contain interaction regions that are geometrically similar to those of seen objects. Aligning these regions with geometrically similar regions on training objects promotes consistent representations, helping the model recognize interaction regions and incorporate their semantic information into the query.
(c) Objects may contain novel interaction structures that cannot be reliably aligned with those observed during training based on geometric similarity. We therefore use a more generalizable 2D affordance model to directly guide feature extraction, facilitating the generation of target-aware queries.
}
\label{motivation}
\end{figure}

Based on this observation, we aim to narrow the semantic gap between queries and target regions. Specifically, we enable the model to reliably capture target-region features and incorporate them into queries, even on unseen objects. First, we encourage geometrically similar interaction regions across training objects to have consistent representations. This allows similar structures on unseen objects to retain familiar feature patterns, making them easier for the model to recognize. Second, novel interaction structures may not be reliably associated with those observed during training through geometric similarity alone. We therefore integrate spatial evidence from a generalizable 2D affordance model to directly guide target-region feature extraction and facilitate the formation of target-aware queries.

To this end, we propose BridgeAfford, which narrows the query–target gap through Interaction Region Alignment (IRA) and Visual Evidence Aggregation (VEA). IRA encourages consistent representations of recurring local interaction structures across objects. We first use Diversity-Aware Initialization (DAI) to construct a compact set of geometrically diverse anchors in a Category-Agnostic Interaction Memory (CAIM). During training, Affinity-Based Contrastive Learning (ABCL) aligns interaction regions with these anchors according to geometric similarity. This encourages the model to represent similar interaction structures consistently despite differences in the objects containing them. When such structures appear on unseen objects, their interaction-region features remain close to those of geometrically similar local structures observed during training.

Complementing this region alignment, VEA provides spatial guidance for target-region feature extraction on the current object. We employ a Normal-Based Gaussian Renderer (NBGR) to generate multi-view renderings and a lightweight 2D affordance model to localize target regions in each view. Confidence-Aware Multi-View Aggregation (CAMVA) then lifts these predictions into dense point-wise 3D visual evidence. We incorporate this evidence as a bias in text-to-point cross-attention, guiding feature extraction toward the target region and yielding more reliable, target-aware queries.

We validate BridgeAfford on the LASO and PIAD datasets. Experimental results demonstrate that our model achieves leading performance, particularly on unseen categories.

In summary, our contributions are as follows:

\begin{itemize}

\item We propose BridgeAfford, a framework that narrows the query--target gap by enabling more reliable target-region feature extraction on unseen objects.

\item We introduce Interaction Region Alignment (IRA) to promote consistent representations of geometrically similar interaction regions across objects, providing transferable cues for target recognition. Complementarily, Visual Evidence Aggregation (VEA) converts multi-view 2D predictions into dense spatial evidence to guide target-region feature extraction and produce target-aware queries.

\item Extensive experiments on LASO and PIAD demonstrate that BridgeAfford outperforms existing methods, particularly in unseen settings, validating its strong generalization ability.

\end{itemize}

\section{Related Work}

\paragraph{3D Affordance Learning.}
3D affordance learning localizes functional regions on 3D objects. Early work formulates it as point-wise multi-label segmentation~\citep{3daffordancenet}, while later methods represent interaction intent using human-object interaction images~\citep{iagnet,great} or language. For language-based grounding, OpenAD~\citep{openad} enables open-vocabulary affordance detection with pretrained text encoders, and LASO~\citep{laso} extends the input to sentence-level instructions. Recent methods~\citep{geal,glance,3daffordancellm,vispla} improve this setting with external models, cross-modal designs, or enhanced training strategies, yet generalization to unseen objects remains challenging.

\paragraph{Object Representation Learning for 3D Affordance.}
LASO~\citep{laso} integrates textual and geometric features through multi-layer upsampling, while GEAL~\citep{geal} improves robustness with 2D--3D consistency alignment. GLANCE~\citep{glance} introduces cross-modal connectors between the 3D and text encoders, and CMAT~\citep{cmat} transfers knowledge from 2D foundation models via cross-modal affinity consistency. Overall, these methods enhance object features by injecting semantic information from 2D or text encoders. Additionally, Aff3DFunc~\citep{aff3dfunc} utilizes contrastive learning to associate point cloud features with functional semantics. \citet{part} employ a set of prototypes to align features corresponding to the same affordance. These methods typically rely on explicit labels to align or distinguish local features. However, in language-guided tasks, interaction intents are implicitly expressed by sentences, making explicit labels unavailable. Extracting such labels with LLMs is costly and does not resolve a deeper issue: the same intent may correspond to geometrically diverse regions across objects, and forcibly aligning them can distort object representations. In contrast to prior works, our method does not rely on explicit affordance labels. Instead, it identifies interaction regions with similar geometric patterns across objects and enforces consistent representations over these regions.

\paragraph{Query Enhancement for 3D Affordance.}
Recent studies have explored different ways to enhance target-aware query representations. 
PART~\citep{part} and HAMMER~\citep{hammer} rely on LLMs or MLLMs to infer interaction-related semantics. 
However, such high-level and abstract semantic priors still remain semantically distant from features of the target region, while also introducing considerable computational overhead. 
In contrast, we employ a lightweight DINOv2-based 2D affordance model to directly predict target-region masks on rendered views and lift them into 3D, providing more explicit spatial guidance.
GLANCE~\citep{glance} further introduces 2D visual cues through sparse keypoint sampling. 
However, sparse samples capture only partial evidence from the target mask and discard its continuous spatial structure and intra-region confidence variation. 
Our CAMVA instead produces dense and smooth point-wise guidance over the entire target region, enabling the query to aggregate more comprehensive regional features while preserving stronger responses at more confident points.

\section{Method}

\subsection{The Backbone of BridgeAfford}

\begin{figure}[t]
    \centering
    \includegraphics[width=1.0\linewidth]{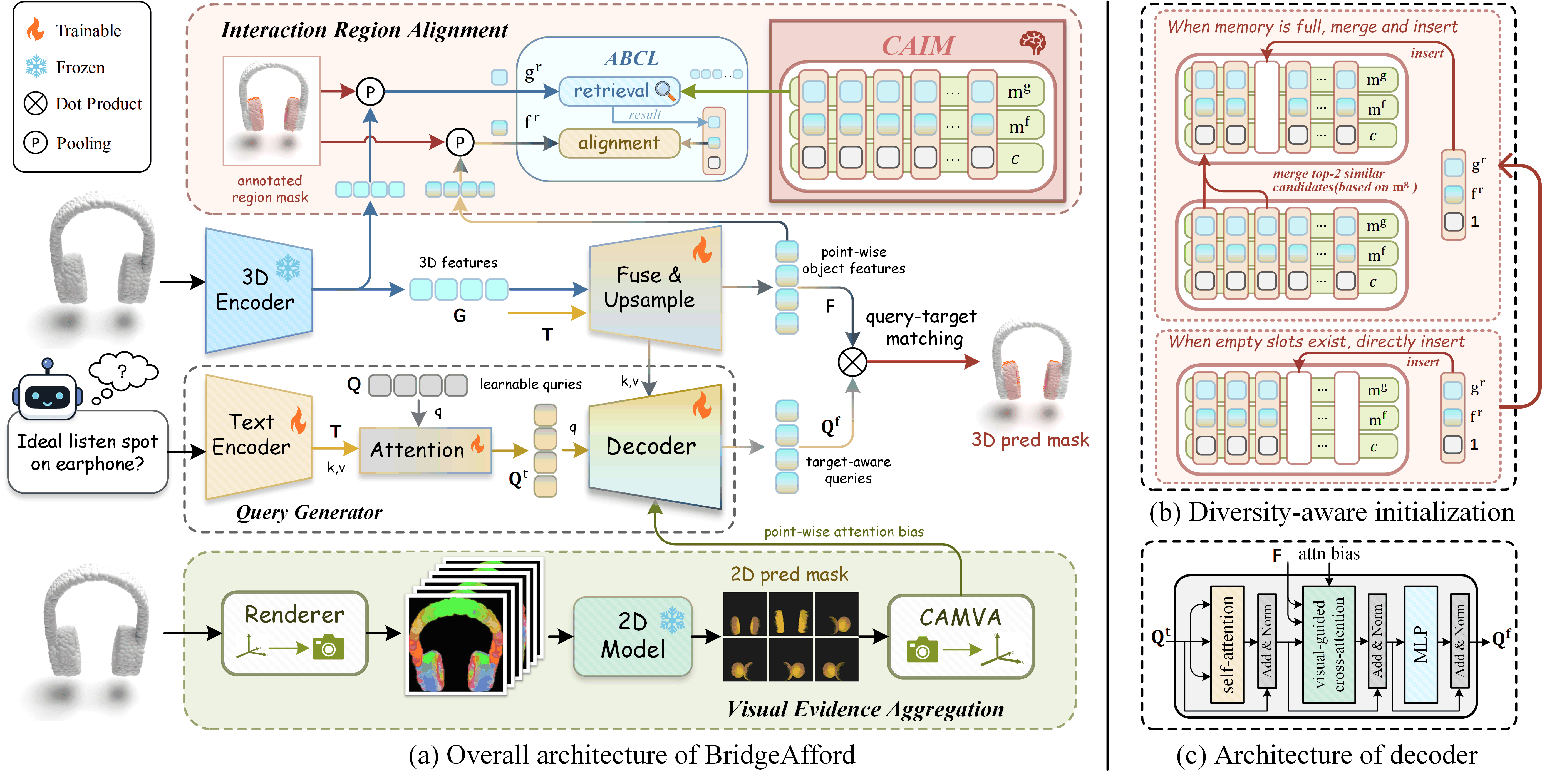}
    \vspace{-1em}
    \caption{
    Overview of BridgeAfford.
    (a) BridgeAfford follows the LASO pipeline and improves generalization from two sides: IRA aligns
geometrically similar interaction regions across objects, while VEA lifts
multi-view 2D affordance predictions into 3D as an attention bias.
    (b) During initialization, new region features are directly inserted when empty slots are available; otherwise, the two most similar entries are merged to release a slot for the new entry.
    (c) Visual evidence is incorporated as an attention bias to guide query-to-point attention toward interaction regions.
    }
    \label{fig:model}
\end{figure}

Given a 3D point cloud $\mathcal{P}=\{p_i\}_{i=1}^{N}$ and an interaction instruction $\mathcal{T}$, language-guided 3D affordance grounding aims to predict a dense affordance map
$\hat{\mathbf{y}}=\{\hat{y}_i\}_{i=1}^{N}$, where $\hat{y}_i\in[0,1]$ denotes the interaction likelihood of point $p_i$. We first extract geometric features $\mathbf{G}$ from the point cloud using a 3D encoder~\citep{uni3d} and fuse them with the text features $\mathbf{T}$ produced by a text encoder~\citep{roberta}. The fused features are then upsampled to obtain point-wise object features $\mathbf{F}$. Meanwhile, in the query generator, a set of learnable queries first extracts semantic information from the text features and subsequently aggregates object representations of the target region through the decoder, yielding target-aware queries $\mathbf{Q}^{f}$. Finally, the affordance mask is obtained by computing the similarity between the target-aware queries and the point-wise object features.

\subsection{Interaction Region Alignment}

\paragraph{Category-Agnostic Interaction Memory (CAIM).}
During training, we extract two region-level features from the annotated interaction region for each sample: a geometric region feature $\mathbf{g}^{r}$ and a language-aware region feature $\mathbf{f}^{r}$. We obtain $\mathbf{g}^{r}$ and $\mathbf{f}^{r}$ by masked average pooling:
\begin{equation}
    \mathbf{g}^{r}
    =
    \operatorname{Norm}
    \left(
    \frac{\sum_{i=1}^{R} y^{g}_{i}\mathbf{g}_{i}}
    {\sum_{i=1}^{R} y^{g}_{i}+\epsilon}
    \right),
    \quad
    \mathbf{f}^{r}
    =
    \operatorname{Norm}
    \left(
    \frac{\sum_{i=1}^{N} y_i\mathbf{f}_{i}}
    {\sum_{i=1}^{N} y_i+\epsilon}
    \right).
\end{equation}
Here, $\mathbf{g}^{r}$ summarizes the local geometric pattern of the interaction region, while $\mathbf{f}^{r}$ summarizes the corresponding object features that integrate both geometric information and interaction semantics. The point-wise mask $\mathbf{y}$ is mapped to group-wise scores $\mathbf{y}^{g}$ according to the group centers of the 3D encoder. CAIM maintains a compact memory bank
\begin{equation}
    \mathcal{M}
    =
    \{(\mathbf{m}^{g}_{k}, \mathbf{m}^{f}_{k}, c_k)\}_{k=1}^{K},
\end{equation}
where each entry stores a geometric region feature $\mathbf{m}^{g}_{k}$, its paired fused feature $\mathbf{m}^{f}_{k}$, and a count $c_k$ that records how many region features have been merged into this entry.

\paragraph{Diversity-Aware Initialization (DAI).}
To capture diverse local geometric structures in the training data, we adopt a diversity-aware strategy to initialize the CAIM. If the
CAIM contains fewer than \(K\) entries, the new feature pair
\((\mathbf{g}^{r},\mathbf{f}^{r})\) is directly inserted with count \(1\).
Once the CAIM is full, we first insert the new feature into the
\((K+1)\)-th temporary slot. CAIM then considers all \(K+1\) entries jointly
and finds the two most similar entries according to their geometric region
features $\mathbf{m}^{g}$:
\begin{equation}
    (u,v)
    =
    \arg\max_{u\neq v,\; u,v\in\{1,\dots,K+1\}}
    \cos(\mathbf{m}^{g}_{u}, \mathbf{m}^{g}_{v}).
\end{equation}
The selected two entries are merged according to their accumulated counts:
\begin{equation}
    \bar{\mathbf{m}}^{g}
    =
    \operatorname{Norm}
    \left(
    \frac{c_u\mathbf{m}^{g}_{u}+c_v\mathbf{m}^{g}_{v}}
    {c_u+c_v}
    \right),
    \quad
    \bar{\mathbf{m}}^{f}
    =
    \operatorname{Norm}
    \left(
    \frac{c_u\mathbf{m}^{f}_{u}+c_v\mathbf{m}^{f}_{v}}
    {c_u+c_v}
    \right),
    \quad
    \bar{c}=c_u+c_v.
\end{equation}
After merging, CAIM is reduced back to \(K\) entries. Repeating this procedure allows CAIM
to maintain \(K\) diverse region features instead of being
dominated by redundant samples.

\paragraph{Affinity-Based Contrastive Learning (ABCL).}
After initialization, we introduce Affinity-Based Contrastive Learning (ABCL) to align $\mathbf{f}^{r}$ based on geometric affinity. Unlike conventional contrastive learning with label-defined positive pairs, ABCL dynamically selects geometrically similar interaction regions as positives, avoiding the forced alignment of regions that share the same interaction function but exhibit substantially different local representations. For each sample, its geometric region feature $\mathbf{g}^{r}$ retrieves the most similar $\mathbf{m}^{g}_{k}$ in CAIM:
\begin{equation}
    k^{+}
    =
    \arg\max_{k}
    \cos(\mathbf{g}^{r},\mathbf{m}^{g}_{k}).
\end{equation}
Here, $\mathbf{m}^{g}_{k^{+}}$ defines the shared geometric pattern for the current interaction region. Its paired feature $\mathbf{m}^{f}_{k^{+}}$ is used as the positive target to align the $\mathbf{f}^{r}$ of the current sample with an InfoNCE loss:
\begin{equation}
    \mathcal{L}_{\mathrm{caim}}
    =
    -
    \log
    \frac{
    \exp(\cos(\mathbf{f}^{r}, \mathbf{m}^{f}_{k^{+}})/\tau)
    }{
    \sum_{k=1}^{K}
    \exp(\cos(\mathbf{f}^{r}, \mathbf{m}^{f}_{k})/\tau)
    }.
\end{equation}
In this way, for each sample, $\mathbf{g}^{r}$ retrieves the most similar interaction region from CAIM through similarity matching, and $\mathbf{f}^{r}$ is aligned with the corresponding $\mathbf{m}^{f}_{k^{+}}$. This encourages the model to produce more general and cross-object-consistent region features. Finally, the selected memory entry is updated online by exponential moving average (EMA):
\begin{equation}
    \mathbf{m}^{g}_{k^{+}}
    \leftarrow
    \operatorname{Norm}
    \left(
    \alpha \mathbf{m}^{g}_{k^{+}} + (1-\alpha)\mathbf{g}^{r}
    \right),
    \quad
    \mathbf{m}^{f}_{k^{+}}
    \leftarrow
    \operatorname{Norm}
    \left(
    \alpha \mathbf{m}^{f}_{k^{+}} + (1-\alpha)\mathbf{f}^{r}
    \right).
\end{equation}

\subsection{Visual Evidence Aggregation}

Geometric alignment may provide insufficient cues for unseen objects with novel or substantially shifted local structures, causing the model to attend to incorrect or incomplete target regions. We therefore introduce Visual Evidence Aggregation (VEA), which converts multi-view 2D affordance predictions into dense 3D visual evidence to guide target-region feature extraction.

\paragraph{Normal-Based Gaussian Renderer (NBGR).}
Renderers such as Mitsuba~3 provide high-quality visualization but are relatively heavy for repeatedly generating multi-view renderings. Following GEAL~\citep{geal}, we adopt Gaussian rendering for efficient multi-view point-cloud rendering. To enrich visual cues for colorless point clouds, we further estimate point normals from local neighborhoods and convert them into RGB colors. This normal-based coloring produces rendered images with clearer surface orientation, object parts, and geometric structures.

\paragraph{2D Affordance Model.}
Given a rendered view and the language instruction, we use a
lightweight 2D affordance model to predict a view-level affordance map. We
instantiate this model with the 2D branch of GEAL~\citep{geal}, which encodes
rendered views with a frozen visual encoder~\citep{dinov2} and fuses visual and
language features for affordance prediction. The model is supervised by 2D
heatmaps rendered from 3D annotations.

\paragraph{Confidence-Aware Multi-View Aggregation (CAMVA).}
For point $p_i$, we project it onto the view $v$ and obtain its 2D affordance score by bilinearly interpolating the values of neighboring pixels. However, direct projection ignores view-dependent visibility: an occluded point may still be projected to an image location, but the sampled 2D score from that view is unreliable. Inspired by D$^{3}$Fields~\citep{d3field}, we compute confidence scores to measure whether the sampled 2D evidence should be trusted. Let $d_i^v$ denote the distance from point $p_i$ to the camera plane in view $v$, and let $\tilde{d}_i^v$ denote the depth value sampled from the rendered depth map. If $p_i$ is visible in view $v$, $\tilde{d}_i^v$ should be close to $d_i^v$. Therefore, we define the depth inconsistency and confidence score as
\begin{equation}
\begin{aligned}
    \Delta_i^v
    &=
    \max
    \left(
    0,
    \frac{d_i^v-\tilde{d}_i^v}{\sigma_d^v+\epsilon}
    \right),
    \quad
    c_i^v
    =
    \mathbb{I}_i^v
    \exp
    \left(
    -\frac{\Delta_i^v}{\tau_d}
    \right),
\end{aligned}
\end{equation}
where $\sigma_d^v$ is the robust depth scale of view $v$, $\mathbb{I}_i^v$ indicates whether $p_i$ has a valid projection and valid depth
in view $v$, and $\tau_d$ controls the depth tolerance. The final 3D visual evidence score is obtained by confidence-weighted aggregation:
\begin{equation}
    s_i
    =
    \frac{
    \sum_{v=1}^{V} c_i^v a_i^v
    }{
    \sum_{v=1}^{V} c_i^v+\epsilon
    }.
\end{equation}
The resulting $s_i$ provides dense point-wise visual evidence in 3D space.

\textbf{Visual-Guided Cross-Attention.}\quad
We inject the aggregated visual evidence into query-to-point attention as an additive bias. For a language-conditioned query $\mathbf{q}^{t}_{m}$ and point representation $\mathbf{f}_{i}$, the attention logit is modified as
\begin{equation}
    \tilde{e}_{m,i}
    =
    \frac{
    (W_q\mathbf{q}^{t}_{m})^{\top}
    (W_k\mathbf{f}_{i})
    }{\sqrt{C}}
    +
    \lambda_{v}s_i,
\end{equation}
where $\lambda_v$ controls the strength of visual guidance. In this way, VEA provides dense spatial evidence that guides language queries toward likely interaction regions.

\subsection{Training Strategy and Objective}

The 2D affordance model is first trained with view-level supervision rendered from 3D point-wise annotations following GEAL~\citep{geal}, and then frozen to produce multi-view affordance heatmaps for VEA. For the 3D model, we optimize the point-wise affordance prediction loss together with the CAIM alignment objective:
\begin{equation}
    \mathcal{L}
    =
    \mathcal{L}_{\mathrm{seg}}
    +
    \lambda_{\mathrm{caim}}\mathcal{L}_{\mathrm{caim}}.
\end{equation}
$\mathcal{L}_{\mathrm{seg}}$ supervises dense affordance scores using a combination of Focal loss and Dice loss, following prior works~\citep{laso,geal,glance}. The CAIM loss encourages geometrically similar interaction regions to share a unified and generalized representation.

Training the 3D backbone proceeds in three stages. First, we train the backbone using only $\mathcal{L}_{\mathrm{seg}}$, so that the text-geometry fusion module and upsampler learn relatively stable object representations. Second, we initialize CAIM with the Diversity-Aware Initialization described above. Third, we jointly optimize $\mathcal{L}_{\mathrm{seg}}$ and $\mathcal{L}_{\mathrm{caim}}$ while updating the CAIM online with EMA. During inference, CAIM is omitted, and BridgeAfford operates with only VEA and the 3D backbone.

\section{Experiments}

\subsection{Experimental Setup}
\paragraph{Implementation Details.}
Following prior work~\citep{glance}, we use Uni3D~\citep{uni3d} as the point-cloud encoder and RoBERTa~\citep{roberta} as the text encoder. The mask decoder is implemented with a Transformer
decoder~\citep{transformer}. We render 6 views for each point cloud at
a resolution of \(224 \times 224\). For the 3D backbone, we train for 30 epochs using AdamW with a batch size
of 16 and a learning rate of \(1\times10^{-4}\). The
CAIM contains \(K=16\) entries, uses 3 warm-up epochs and 1
initialization epoch, and is then updated online with EMA coefficient 0.999.
In all experiments, the 2D model and the 3D backbone are trained using the same training data. All experiments were conducted on a single NVIDIA GeForce RTX 5090 GPU.
Detailed hyperparameter configurations are presented in Appendix~\ref{app:hyperparameters}.

\paragraph{Datasets.}
We evaluate BridgeAfford on the LASO~\citep{laso} and PIAD~\citep{iagnet} datasets. LASO is a language-guided 3D affordance
segmentation benchmark. To evaluate PIAD in the language-guided
setting, we directly reuse the PIAD language templates provided by
GEAL~\citep{geal}. Both datasets are evaluated under seen and unseen settings. In the seen
setting, training and testing share the same object categories or
object-affordance compositions. In the unseen setting, specific
object-affordance pairs are held out during training and appear only during
testing. Therefore, unseen performance directly reflects the model's ability
to generalize beyond memorized object-region associations, which is the main
focus of our work.

\paragraph{Evaluation Metrics.}
Following prior work, we report average Intersection over Union (aIoU), Area
Under the ROC Curve (AUC), Similarity (SIM), and Mean Absolute Error (MAE). Higher aIoU, AUC,
and SIM indicate better performance, while lower MAE is preferred.

\subsection{Quantitative Results}

\paragraph{Results on LASO.}
Table~\ref{tab:laso-main} compares BridgeAfford with existing methods on LASO.
On seen objects, BridgeAfford achieves strong performance and remains competitive
with recent approaches. More importantly, BridgeAfford achieves leading performance across all metrics in the unseen setting, demonstrating its strong generalization ability to unseen objects.

\begin{table}[H]
\caption{
Results on the LASO dataset.
Higher is better for aIoU, AUC, and SIM, while lower is better for MAE.
The best and second-best results are highlighted in bold and underlined,
respectively.
}
\label{tab:laso-main}
\centering
\resizebox{\linewidth}{!}{
\renewcommand{\arraystretch}{1.12}
\begin{tabular}{llcccccccc}
\toprule
& & \multicolumn{4}{c}{Seen} & \multicolumn{4}{c}{Unseen} \\
\cmidrule(lr){3-6}\cmidrule(lr){7-10}
Method & Venue
& aIoU$\uparrow$ & AUC$\uparrow$ & SIM$\uparrow$ & MAE$\downarrow$
& aIoU$\uparrow$ & AUC$\uparrow$ & SIM$\uparrow$ & MAE$\downarrow$ \\
\midrule
ReferTrans~\citep{refertrans}
& NeurIPS 2021
& 13.7 & 79.8 & 0.497 & 0.124
& 10.2 & 69.1 & 0.432 & 0.145 \\
ReLA~\citep{rela}
& CVPR 2023
& 15.2 & 78.9 & 0.532 & 0.118
& 10.7 & 69.7 & 0.429 & 0.144 \\
3D-SPS~\citep{3d-sps}
& CVPR 2022
& 11.4 & 76.2 & 0.433 & 0.138
& 7.9 & 68.8 & 0.402 & 0.158 \\
IAGNet~\citep{iagnet}
& ICCV 2023
& 17.8 & 82.3 & 0.561 & 0.109
& 12.9 & 77.8 & 0.443 & 0.129 \\
PointRefer~\citep{laso}
& CVPR 2024
& 20.8 & \underline{87.3} & 0.629 & 0.093
& 14.6 & 80.2 & 0.507 & 0.119 \\
3D-AffLLM~\citep{3daffordancellm}
& ICLR 2025
& 18.2 & 84.9 & 0.622 & 0.104
& 15.3 & 78.7 & 0.542 & 0.124 \\
GEAL~\citep{geal}
& CVPR 2025
& 22.0 & 86.7 & 0.634 & 0.092
& 16.7 & 80.9 & 0.567 & 0.106 \\
ViSPLA~\citep{vispla}
& NeurIPS 2025
& 22.8 & \underline{87.3} & \textbf{0.651} & 0.090
& 17.1 & 81.5 & \underline{0.571} & \underline{0.103} \\
GLANCE~\citep{glance}
& ICCV 2025
& \underline{23.3} & \textbf{88.1} & 0.632 & \underline{0.089}
& \underline{18.7} & \underline{81.7} & 0.532 & 0.112 \\
\rowcolor{oursblue}
\textbf{BridgeAfford (Ours)}
& --
& \textbf{23.7} & 86.8 & \underline{0.646} & \textbf{0.086}
& \textbf{21.8} & \textbf{83.0} & \textbf{0.597} & \textbf{0.099} \\
\bottomrule
\end{tabular}
}
\end{table}
\vspace{-1em}

\paragraph{Results on PIAD.}
Table~\ref{tab:piad-main} reports the results on PIAD. Among language-guided methods, BridgeAfford achieves
the best performance under both seen and unseen settings. Compared with ViSPLA, which introduces an LLM for affordance
reasoning, BridgeAfford still achieves clear gains, improving aIoU from
9.20 to 14.50. In addition, we compare BridgeAfford with current
intention-driven methods, and the results show that our method still achieves
the best performance on most metrics.

\begin{table}[H]
\caption{
Results on the PIAD dataset.
The best and second-best results among all methods are highlighted in bold
and underlined, respectively.
}
\label{tab:piad-main}
\centering
\resizebox{\linewidth}{!}{
\renewcommand{\arraystretch}{1.12}
\begin{tabular}{llcccccccc}
\toprule
& & \multicolumn{4}{c}{Seen} & \multicolumn{4}{c}{Unseen} \\
\cmidrule(lr){3-6}\cmidrule(lr){7-10}
Method & Venue
& aIoU$\uparrow$ & AUC$\uparrow$ & SIM$\uparrow$ & MAE$\downarrow$
& aIoU$\uparrow$ & AUC$\uparrow$ & SIM$\uparrow$ & MAE$\downarrow$ \\
\midrule
\multicolumn{10}{c}{\textit{Intention-Driven 3D Affordance Grounding}} \\
\midrule
PMF~\citep{pmf}
& ICCV 2021
& 10.13 & 75.05 & 0.425 & 0.141
& 4.67 & 60.25 & 0.330 & 0.211 \\

ILN~\citep{iln}
& SIGGRAPH 2022
& 11.25 & 75.84 & 0.427 & 0.137
& 4.71 & 59.69 & 0.325 & 0.207 \\

FRCNN~\citep{frcnn}
& Remote Sensing 2023
& 11.97 & 76.05 & 0.429 & 0.136
& 4.71 & 61.92 & 0.332 & 0.195 \\

PFusion~\citep{pfusion}
& CVPR 2018
& 12.31 & 77.50 & 0.432 & 0.135
& 5.33 & 61.87 & 0.330 & 0.193 \\

XMFNet~\citep{xmfnet}
& NeurIPS 2022
& 12.94 & 78.25 & 0.441 & 0.127
& 5.68 & 62.58 & 0.342 & 0.188 \\

IAGNet~\citep{iagnet}
& ICCV 2023
& 20.51 & 84.85 & 0.545 & 0.098
& 7.95 & 71.84 & 0.352 & 0.127 \\

GREAT~\citep{great}
& CVPR 2025
& 19.61 & 85.22 & 0.569 & 0.093
& 8.32 & 67.46 & 0.330 & 0.121 \\

HAMMER~\citep{hammer}
& CVPR 2026
& 22.20 & \textbf{88.43} & 0.605 & \underline{0.083}
& \underline{13.71} & \textbf{80.92}
& \underline{0.449} & 0.109 \\

\midrule
\multicolumn{10}{c}{\textit{Language-Guided 3D Affordance Grounding}} \\
\midrule
PointRefer~\citep{laso}
& CVPR 2024
& 19.70 & 84.20 & 0.590 & 0.096
& 8.00 & 69.20 & 0.386 & 0.118 \\

GEAL~\citep{geal}
& CVPR 2025
& 22.50 & 85.00 & 0.600 & 0.092
& 8.70 & 72.50 & 0.390 & 0.102 \\

ViSPLA~\citep{vispla}
& NeurIPS 2025
& \underline{23.10} & 85.80 & \underline{0.664} & 0.089
& 9.20 & 73.10 & 0.431 & \underline{0.099} \\

\rowcolor{oursblue}
\textbf{BridgeAfford (Ours)}
& --
& \textbf{25.40} & \underline{87.60}
& \textbf{0.668} & \textbf{0.080}
& \textbf{14.50} & \underline{78.80}
& \textbf{0.459} & \textbf{0.095} \\

\bottomrule
\end{tabular}
}
\end{table}

\newpage
\paragraph{Cross-Dataset Evaluation.}
To evaluate cross\mbox{-}dataset transferability, we train models on the
LASO dataset and directly evaluate them on the PIAD test set. As shown in
Table~\ref{tab:cross-dataset}, our method achieves the best performance
across all reported metrics. The results of PointRefer and GEAL are from
our reproduction, while the results of 3D-AffLLM and ViSPLA are taken from
ViSPLA~\citep{vispla}. Note that ViSPLA does not report MAE, and therefore
the corresponding entry is unavailable.

\begin{table}[htbp]
\centering
\caption{Cross-dataset generalization.}
\label{tab:cross-dataset}
\renewcommand{\arraystretch}{1.10}
\begin{tabular}{lcccc}
\toprule
Method
& aIoU$\uparrow$ & AUC$\uparrow$ & SIM$\uparrow$ & MAE$\downarrow$ \\
\midrule
PointRefer~\citep{laso}
& 17.7 & 83.2 & 0.567 & 0.110 \\
GEAL~\citep{geal}
& 18.5 & 83.8 & 0.582 & 0.096 \\
3D-AffLLM~\citep{3daffordancellm}
& 17.6 & 82.4 & 0.570 & -- \\
ViSPLA~\citep{vispla}
& 19.7 & 84.5 & 0.610 & -- \\
Ours
& \textbf{23.6} & \textbf{87.0} & \textbf{0.633} & \textbf{0.080} \\
\bottomrule
\end{tabular}
\end{table}

\begin{figure}[H]
    \centering
    \includegraphics[width=1.0\linewidth]{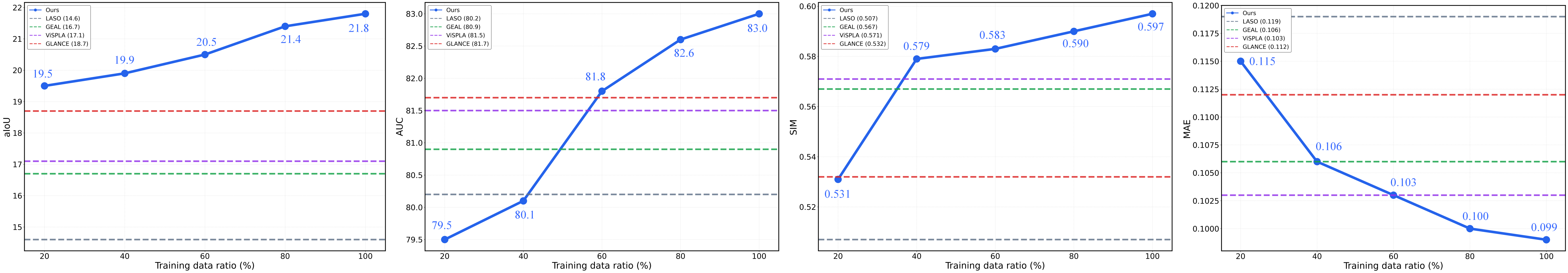}
    \caption{Data efficiency of our method with limited training data.}
    \label{fig:data_efficiency}
\end{figure}

\paragraph{Generalization with Limited Training Data.}
To evaluate the data efficiency of BridgeAfford, we randomly retain $\{20\%,40\%,60\%,80\%,100\%\}$ of the original LASO unseen training split, while ensuring a balanced sampling distribution across different object–affordance combinations. As shown in Figure~\ref{fig:data_efficiency}, the blue curves denote the performance of BridgeAfford under different training data ratios, while the horizontal lines indicate existing baselines trained with the full training set. Notably, BridgeAfford trained with only 60\% of the data surpasses all baselines trained with 100\% of the data, and achieves the best aIoU with only 20\% of the data. These results demonstrate the superior data efficiency and strong generalization of BridgeAfford under limited training data.
\vspace{-1em}
\subsection{Ablation Studies}
\label{sec:ablation}
\paragraph{Main Module Ablation.}
We evaluate the contribution of the two main components on the LASO dataset.
As shown in Table~\ref{tab:ablation-main-module}, both IRA and VEA consistently
improve the backbone, especially in the unseen setting. IRA improves the
generalization by encouraging geometrically similar
interaction regions across objects to learn consistent representations. VEA
provides dense target-region evidence to guide queries
toward target regions. IRA improves the transferability of interaction-region representations, whereas VEA provides dense spatial evidence for target-region feature extraction. Combining them therefore achieves the best overall performance.

\begin{table}[H]
\caption{
Main module ablation on the LASO dataset.
}
\label{tab:ablation-main-module}
\centering
\resizebox{0.9\linewidth}{!}{
\renewcommand{\arraystretch}{1.12}
\begin{tabular}{lcccccccc}
\toprule
& \multicolumn{4}{c}{Seen} & \multicolumn{4}{c}{Unseen} \\
\cmidrule(lr){2-5}\cmidrule(lr){6-9}
Variant
& aIoU$\uparrow$ & AUC$\uparrow$ & SIM$\uparrow$ & MAE$\downarrow$
& aIoU$\uparrow$ & AUC$\uparrow$ & SIM$\uparrow$ & MAE$\downarrow$ \\
\midrule
Backbone
& 22.0 & 84.6 & 0.610 & 0.094
& 19.3 & 80.7 & 0.568 & 0.107 \\
Backbone + IRA
& 23.4 & 85.7 & 0.635 & 0.090
& 20.7 & 82.5 & 0.585 & 0.100 \\
Backbone + VEA
& 22.9 & 86.0 & 0.638 & 0.089
& 20.4 & 81.1 & 0.592 & 0.102 \\
Backbone + IRA + VEA
& \textbf{23.7} & \textbf{86.8} & \textbf{0.646} & \textbf{0.086}
& \textbf{21.8} & \textbf{83.0} & \textbf{0.597} & \textbf{0.099} \\
\bottomrule
\end{tabular}
}
\end{table}

\paragraph{Label-Based Alignment vs. Geometry-Based Alignment.}
We evaluate an alternative strategy that aligns interaction regions across objects according to affordance labels. As shown in Table~\ref{tab:ablation-fine-grained}, label-based alignment underperforms geometry-based alignment. Since regions with the same label may exhibit substantially different geometries, enforcing their alignment can distort the original object representations.

\paragraph{Random Initialization vs. Diversity-Aware Initialization (DAI).} 
To validate the role of DAI, we randomly initialize $K$ vectors, which are then updated during training. The results show that this variant improves over the backbone, but underperforms the DAI strategy. Notably, we observe that random initialization can easily lead to training collapse, preventing the model from learning meaningful representations. This is mainly because, at the early training stage, local features may be aligned to uninformative entries, thereby corrupting the representation learning process.

\paragraph{2D Affordance Model Only.}
We also evaluate a variant that directly uses the point-wise scores obtained by aggregating multi-view 2D predictions into 3D as the final output, without using the 3D model. The results in Table~\ref{tab:ablation-fine-grained} show that this variant performs substantially worse than the full 3D model. This is mainly because limited viewpoints and occlusions prevent 2D predictions from forming geometrically continuous and smooth masks in 3D. In Appendix~\ref{app:2d-vs-3d}, we further provide empirical evidence that 2D predictions exhibit stronger generalization in localizing target regions, but remain insufficient for producing high-quality point-wise masks.

\paragraph{Without Normal-Based Coloring.}
We remove the normal-based coloring mechanism and render point clouds following the rendering strategy of GEAL~\citep{geal}. Results in Table~\ref{tab:ablation-fine-grained} show that this variant leads to a performance drop. This is mainly because different surfaces and parts of the same object lack color distinction, making the object and its components harder to recognize. In Appendix~\ref{app:rendering}, we further illustrate the rendering results of our method, demonstrating that normal-based coloring significantly enhances the visual quality of the rendered images.

\paragraph{Without Depth Confidence.}
We remove the confidence scores computed from depth consistency and projection validity, and directly average the sampled 2D scores of each 3D point across all views as its visual evidence. The results show that removing depth confidence leads to performance degradation because occlusion is no longer properly accounted for during multi-view aggregation. Consequently, the 2D masks cannot be accurately lifted into point-wise 3D visual evidence.

\begin{table}[t]
\caption{Fine-grained ablation on the LASO unseen split.}
\label{tab:ablation-fine-grained}
\centering
\setlength{\tabcolsep}{3pt}
\renewcommand{\arraystretch}{1.35}
\resizebox{\linewidth}{!}{%
\begin{tabular}{lcccc|lcccc}
\hline
\multicolumn{5}{c|}{\textit{IRA fine-grained ablation}} &
\multicolumn{5}{c}{\textit{VEA fine-grained ablation}} \\
\hline
Variant & aIoU$\uparrow$ & AUC$\uparrow$ & SIM$\uparrow$ & MAE$\downarrow$ &
Variant & aIoU$\uparrow$ & AUC$\uparrow$ & SIM$\uparrow$ & MAE$\downarrow$ \\
\hline
Label-based alignment & 20.1 & 80.5 & 0.563 & 0.102 &
2D affordance model only & 15.2 & 77.1 & 0.500 & 0.192 \\
Random initialization & 19.8 & 81.4 & 0.588 & 0.102 &
w/o normal-based coloring & 19.1 & 81.0 & 0.565 & 0.108 \\
Full IRA & \textbf{21.8} & \textbf{83.0} & \textbf{0.597} & \textbf{0.099} &
w/o depth confidence & 21.4 & 81.5 & 0.592 & 0.106 \\
& & & & &
Full VEA & \textbf{21.8} & \textbf{83.0} & \textbf{0.597} & \textbf{0.099} \\
\hline
\end{tabular}%
}
\end{table}

\subsection{Qualitative Results}
\begin{figure}[H]
    \centering
    \includegraphics[width=1.0\linewidth]{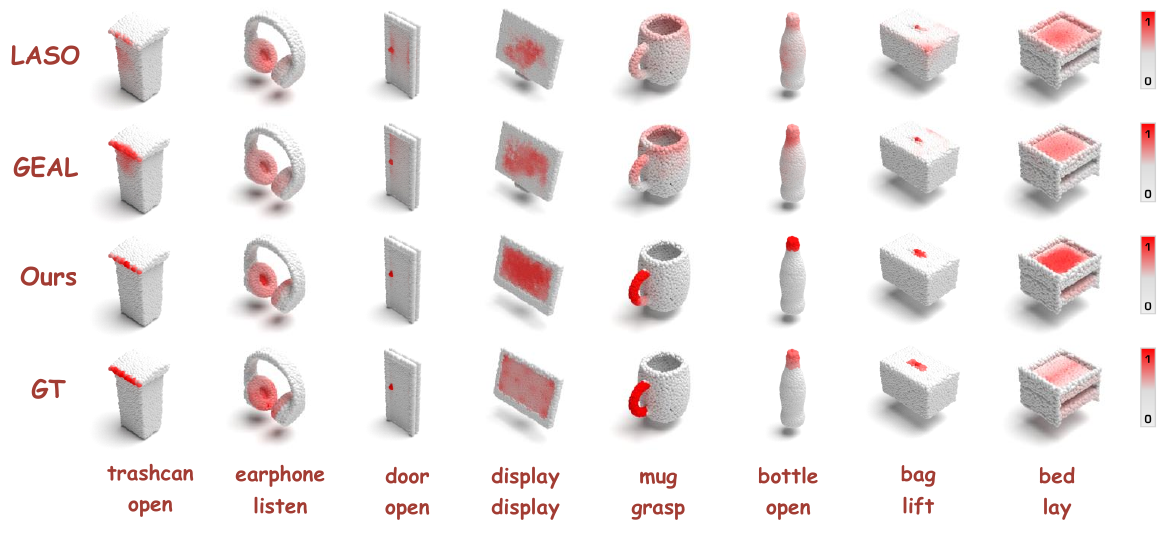}
    \vspace{-1em}
    \caption{Qualitative comparisons on the LASO dataset (unseen).}
    \label{fig:qualitative}
\end{figure}

As shown in Figure~\ref{fig:qualitative}, we visualize representative cases on the LASO unseen split and compare BridgeAfford with previous methods. Our method localizes the target interaction regions more accurately, produces predictions with higher confidence, and effectively suppresses noisy responses in irrelevant regions, resulting in cleaner and more complete affordance masks. These qualitative results further demonstrate the superior generalization ability of our method on unseen objects.

\vspace{-1em}

\section{Conclusion}
BridgeAfford improves 3D affordance grounding on unseen objects by narrowing the semantic gap between queries and target regions. Our results suggest that transferable local geometric structures learned from similar interaction regions contribute to cross-object generalization, while 2D visual evidence provides valuable guidance for query localization. Extensive experiments on LASO and PIAD validate the superiority of BridgeAfford, highlighting the potential of combining geometric structures with 2D visual priors for generalizable 3D affordance understanding.

\bibliography{references}
\bibliographystyle{preprint}

\appendix
\clearpage

\section{Hyperparameter Analysis}
\label{app:hyperparameters}

We analyze the key hyperparameters of Interaction Region Alignment and Visual Evidence Aggregation. Specifically, we investigate the memory size \(K\), the number of warm-up epochs, the CAIM loss weight \(\lambda_{\mathrm{caim}}\), the EMA coefficient, the visual attention bias coefficient \(\lambda_v\), and the number of rendered views. All hyperparameters are selected on the LASO unseen validation split, while the main results reported in the paper are evaluated on the corresponding test split.

\begin{table}[H]
\caption{Hyperparameter selection of IRA-related components on the LASO
unseen split.}
\label{tab:app-hyper-caim}
\centering
\small
\setlength{\tabcolsep}{7pt}
\renewcommand{\arraystretch}{1.08}
\resizebox{0.9\linewidth}{!}{%
\begin{tabular}{lccccc}
\toprule
Hyperparameter & Value & aIoU$\uparrow$ & AUC$\uparrow$ & SIM$\uparrow$ & MAE$\downarrow$ \\
\midrule
Memory size $K$ & 4  & 20.7 & 82.0 & 0.583 & 0.104 \\
Memory size $K$ & 8  & 21.5 & 82.6 & 0.588 & 0.102 \\
Memory size $K$ & \textbf{16} & \textbf{22.0} & \textbf{83.9} & \textbf{0.598} & \textbf{0.096} \\
Memory size $K$ & 32 & 21.5 & 83.1 & 0.596 & 0.101 \\
\midrule
Warm-up epochs & 0  & 20.8 & 82.7 & 0.582 & 0.102 \\
Warm-up epochs & 1  & 21.0 & 83.4 & 0.594 & 0.099 \\
Warm-up epochs & \textbf{3} & \textbf{22.0} & \textbf{83.9} & \textbf{0.598} & \textbf{0.096} \\
Warm-up epochs & 5  & 21.5 & 82.5 & 0.592 & 0.101 \\
\midrule
Cross-object alignment $\lambda_{\mathrm{caim}}$
& 0.1 & 20.4 & 82.7 & 0.585 & 0.106 \\
Cross-object alignment $\lambda_{\mathrm{caim}}$
& 0.5  & 20.9 & 83.0 & 0.597 & 0.104 \\
Cross-object alignment $\lambda_{\mathrm{caim}}$
& \textbf{1.0} & \textbf{22.0} & \textbf{83.9} & \textbf{0.598} & \textbf{0.096} \\
Cross-object alignment $\lambda_{\mathrm{caim}}$
& 1.5  & 21.5 & 83.7 & 0.586 & 0.099 \\
\midrule
EMA coefficient
& 0.9 & 21.0 & 83.3 & 0.590 & 0.101 \\
EMA coefficient
& 0.99 & 21.4 & 83.1 & 0.595 & 0.102 \\
EMA coefficient
& \textbf{0.999} & \textbf{22.0} & \textbf{83.9} & \textbf{0.598} & \textbf{0.096} \\
\bottomrule
\end{tabular}
}
\end{table}

\begin{table}[H]
\caption{Hyperparameter selection of VEA-related components on the LASO
unseen split.}
\label{tab:app-hyper-vea}
\centering
\small
\setlength{\tabcolsep}{7pt}
\renewcommand{\arraystretch}{1.08}
\resizebox{0.9\linewidth}{!}{%
\begin{tabular}{lccccc}
\toprule
Hyperparameter & Value & aIoU$\uparrow$ & AUC$\uparrow$ & SIM$\uparrow$ & MAE$\downarrow$ \\
\midrule
Attention bias $\lambda_v$ & 0.1 & 20.7 & 81.5 & 0.584 & 0.109 \\
Attention bias $\lambda_v$ & 0.5  & 21.6 & 82.8 & 0.594 & 0.100 \\
Attention bias $\lambda_v$ & \textbf{1.0} & \textbf{22.0} & \textbf{83.9} & \textbf{0.598} & \textbf{0.096} \\
Attention bias $\lambda_v$ & 2.0  & 21.3 & 82.1 & 0.589 & 0.103 \\
\midrule
Number of views & 2  & 21.2 & 82.9 & 0.589 & 0.100 \\
Number of views & 4  & 21.7 & 83.4 & 0.596 & 0.098 \\
Number of views & \textbf{6} & \textbf{22.0} & \textbf{83.9} & 0.598 & \textbf{0.096} \\
Number of views & 8 & 22.0 & 83.7 & \textbf{0.601 }& 0.098 \\

\bottomrule
\end{tabular}
}
\end{table}

\section{Quantitative Analysis of the Query–Target Gap}
In this section, we quantify the query–target gap and examine the effects of individual modules from this perspective. As shown in Table~\ref{tab:query_target_similarity}, we measure the similarity between queries and target-region features under four configurations. Specifically, we first compute a target-region representation by averaging point-wise features weighted by the ground-truth affordance scores. We then calculate the cosine similarity between each query feature and this pooled representation and average the similarities across queries to obtain the query–target similarity. For comparison, we also measure the similarity between queries and non-target-region features. The results show that both IRA and VEA improve query–target similarity, with their combination achieving the highest target-region similarity.

\begin{table}[t]
    \centering
    \caption{Query--region cosine similarity on the evaluation set.
    $\Delta$ denotes target similarity minus background similarity.}
    \label{tab:query_target_similarity}
    \begin{tabular}{lccc}
        \toprule
        Method & Target $\uparrow$ & Background $\downarrow$
               & $\Delta$ $\uparrow$ \\
        \midrule
        Backbone       & $-0.009$        & $-0.202$        & $0.193$ \\
        Backbone + IRA & $ 0.045$        & $-0.161$        & $0.206$ \\
        Backbone + VEA & $ 0.021$        & $\mathbf{-0.204}$        & $0.225$ \\
        Backbone + IRA + VEA    & $\textbf{0.062}$ & $-0.183$        & $\textbf{0.245}$ \\
        \bottomrule
    \end{tabular}
\end{table}

\section{Visual Analysis of Diversity-Aware Initialization}
\label{app:dai}
\begin{figure}[H]
    \centering
    \includegraphics[width=0.5\linewidth]{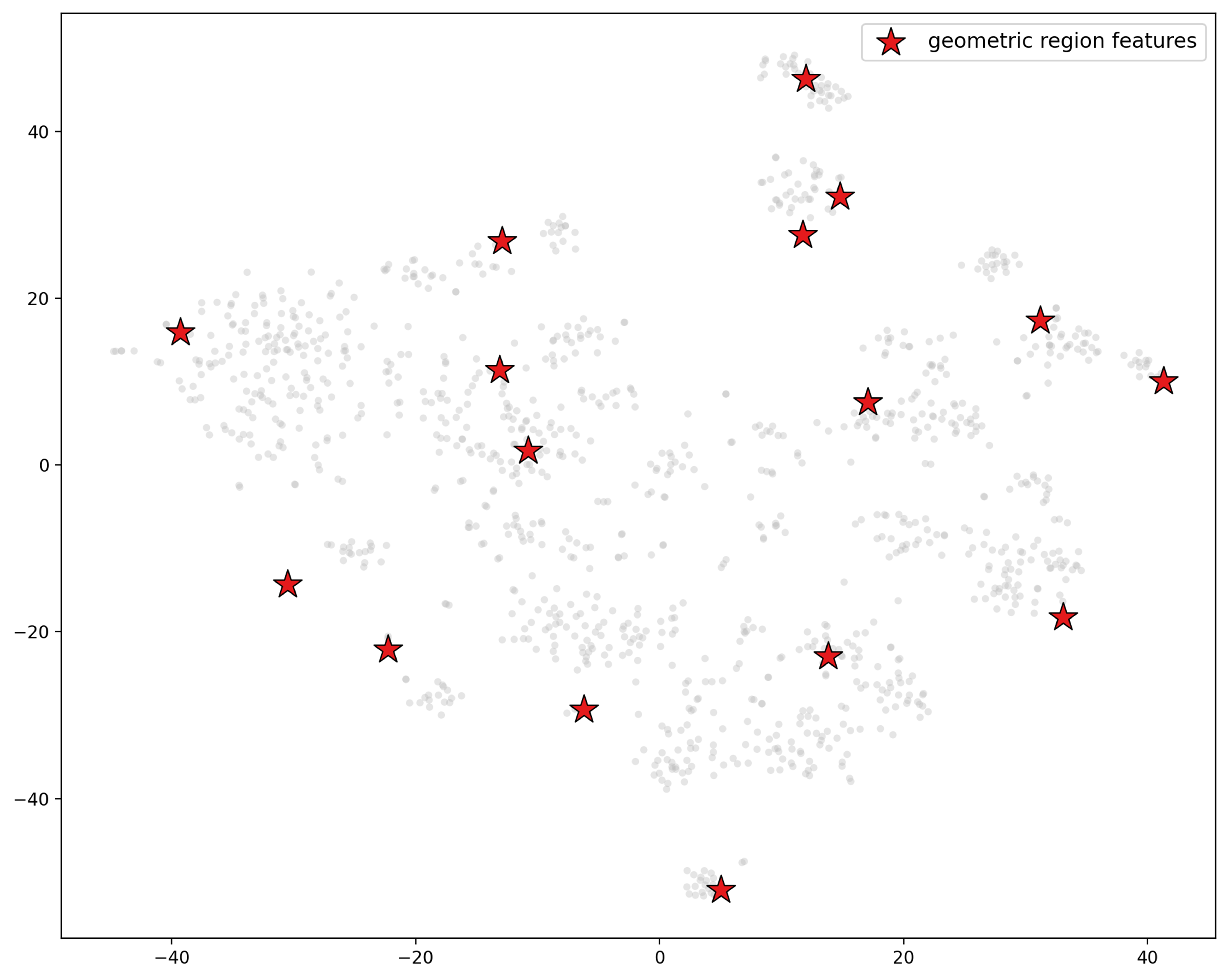}
    \caption{t-SNE visualization of geometric region features after DAI
    initialization.}
    \label{fig:dai}
\end{figure}

To inspect whether DAI preserves diverse geometric patterns, we visualize geometric region features of each entry in the initialized CAIM using t-SNE. As shown in Figure~\ref{fig:dai}, the red stars denote the 16 geometric features, while the gray points represent the target-region geometric features extracted from a subset of samples randomly selected from the dataset with balanced sampling across object categories. The 16 stars are widely distributed in the feature space, indicating that DAI effectively captures diverse geometric patterns.  

\section{Further Analysis of Visual Evidence Aggregation}

\subsection{Advantages of Normal-Based Gaussian Renderer}
\label{app:rendering}
\begin{figure}[H]
\centering
\includegraphics[width=\linewidth]{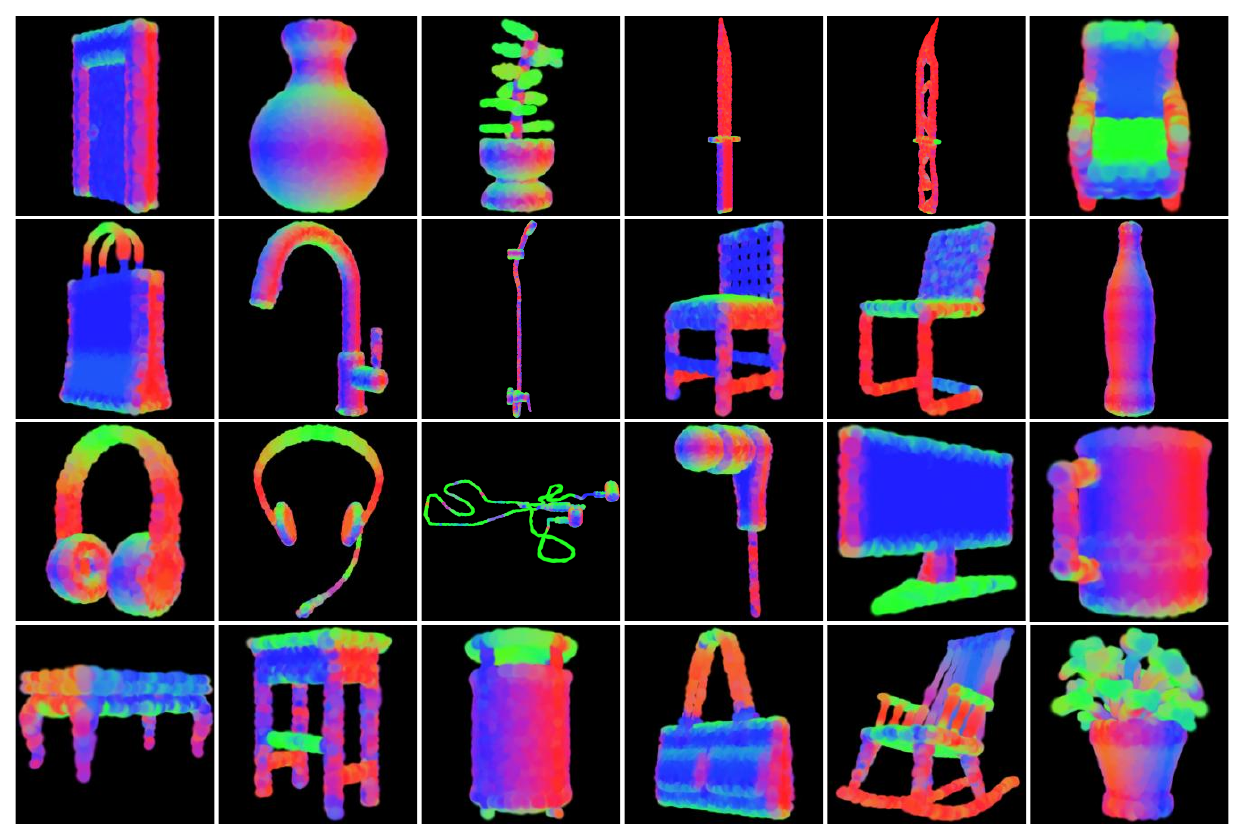}
\vspace{-2mm}
\caption{
Visualization of renderings generated by Normal-Based Gaussian Renderer.
}
\label{fig:rendering}
\end{figure}

We qualitatively demonstrate that NBGR can generate more informative rendered images for the 2D affordance model. As illustrated in Figure~\ref{fig:rendering}, NBGR produces geometry-aware renderings, where different surface regions exhibit distinguishable appearances and object components preserve clear geometric boundaries. Such renderings provide richer spatial cues for subsequent affordance prediction.

\subsection{Generalization Capability of the 2D Affordance Model versus the 3D Backbone}
\label{app:2d-vs-3d}
As discussed in Section~\ref{sec:ablation}, directly adopting aggregated 2D masks as the final output for 3D affordance grounding leads to performance degradation. However, this does not necessarily indicate weaker generalization of 2D models. To investigate this, we project the predictions of the pure 3D backbone onto the image plane and compare them with the 2D affordance predictions. Since the projection shares the same rendering process used for data generation, the projection error is negligible, enabling a fair comparison between 2D and 3D models. As shown in Table~\ref{tab:2d-vs-3d}, the 2D affordance model achieves better performance, highlighting its strong generalization ability and effectiveness as a visual prior for 3D affordance grounding.

\begin{table}[H]
\caption{Comparison of the generalization capabilities of the 2D affordance model and the 3D backbone on the LASO unseen split.}
\label{tab:2d-vs-3d}
\centering
\small
\setlength{\tabcolsep}{8pt}
\renewcommand{\arraystretch}{1.08}
\begin{tabular}{lcccc}
\toprule
Method & aIoU$\uparrow$ & AUC$\uparrow$ & SIM$\uparrow$ & MAE$\downarrow$ \\
\midrule
3D Backbone (Projected) & 21.7 & 91.9 & 0.645 & 0.042 \\
2D Affordance Model & 25.2 & 92.2 & 0.704 & 0.040 \\
\bottomrule
\end{tabular}
\end{table}

\subsection{Dense Point-wise Evidence vs. Sparse Keypoint Sampling}
In this section, we compare the sparse keypoint sampling strategy introduced by GLANCE with our CAMVA to investigate how different visual guidance strategies convert multi-view 2D affordance masks into effective 3D guidance. For a fair comparison, both variants use the same 3D backbone, 2D affordance model, multi-view predictions, and training settings, differing only in how the 2D visual evidence is aggregated and incorporated into the 3D model. The GLANCE-style strategy selects a sparse set of reliable 3D keypoints from the multi-view masks as query priors, whereas CAMVA performs confidence-weighted multi-view aggregation based on projection validity and depth consistency, producing dense point-wise visual evidence that is incorporated as a cross-attention bias.

As shown in Table~\ref{tab:guidance_comparison}, CAMVA consistently outperforms the sparse keypoint strategy across all metrics. These results indicate that preserving and aggregating dense point-wise confidence information provides more effective spatial guidance for target-region feature aggregation than representing the multi-view masks with a sparse set of keypoints.

\begin{table}[H]
\centering
\caption{
Comparison of different visual guidance strategies on the LASO unseen split.
All variants use the same backbone, 2D affordance model, multi-view predictions,
and training settings.
}
\label{tab:guidance_comparison}
\resizebox{\columnwidth}{!}{
\begin{tabular}{lccccc}
\toprule
Guidance Strategy
& Guidance Form
& aIoU $\uparrow$
& AUC $\uparrow$
& SIM $\uparrow$
& MAE $\downarrow$
\\
\midrule
GLANCE-style
& Sparse keypoint guidance
& 21.2 & 81.9 & 0.585 &  0.103   \\
CAMVA (Ours)
& Dense point-wise evidence
& \textbf{21.8} & \textbf{83.0 } & \textbf{0.597 }&\textbf{ 0.099}  \\
\bottomrule
\end{tabular}
}
\end{table}

\subsection{Effect of Visual Evidence Aggregation on Query Attention}

\begin{figure}[H]
\centering
\includegraphics[width=\linewidth]{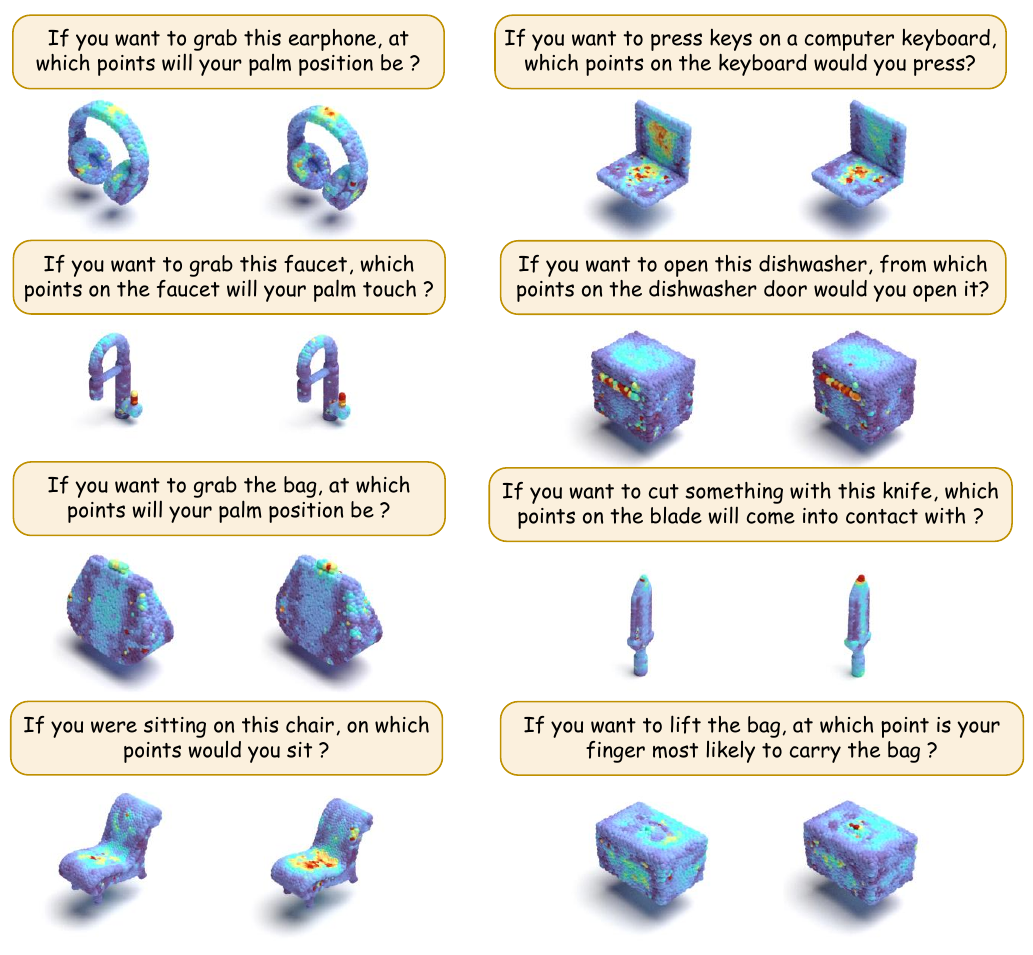}
\vspace{-2mm}

\caption{Attention visualization with and without visual guidance.}

\label{fig:attn_vis}
\end{figure}

To further verify that VEA helps improve the attention distribution of queries,
we visualize the attention maps of queries with and without visual evidence
guidance. As shown in Figure~\ref{fig:attn_vis}, without the additional 2D visual evidence, queries tend to
attend to irrelevant regions on unseen objects or fail to capture the complete
interaction areas. In contrast, after incorporating visual evidence through
VEA, the attention distributions become more concentrated on the target regions, while suppressing responses from irrelevant background
areas.

\newpage
\section{Fine-Grained Results by object–affordance combination}
\label{app:fine-grained}
To provide a more detailed evaluation, Table \ref{tab:laso_unseen_per_class} reports the performance of BridgeAfford on each object–affordance combination in the LASO unseen split across four metrics: aIoU, AUC, SIM, and MAE.

\begin{table}[H]
\centering
\caption{Per-class results on the LASO unseen split.}
\label{tab:laso_unseen_per_class}
\resizebox{\textwidth}{!}{
\begin{tabular}{lcccc|lcccc}
\toprule
Class & aIoU & AUC & SIM & MAE
& Class & aIoU & AUC & SIM & MAE \\
\midrule
bag-contain & 29.40 & 98.21 & 0.796 & 0.092
& bag-grasp & 20.86 & 83.71 & 0.507 & 0.106 \\

bag-lift & 35.39 & 89.36 & 0.538 & 0.026
& bag-open & 13.90 & 83.21 & 0.523 & 0.077 \\

bed-lay & 25.38 & 86.16 & 0.625 & 0.103
& bed-sit & 13.59 & 85.17 & 0.521 & 0.095 \\

bed-support & 13.69 & 67.55 & 0.495 & 0.127
& bottle-contain & 25.42 & 87.56 & 0.570 & 0.122 \\

bottle-grasp & 26.25 & 81.23 & 0.572 & 0.140
& bottle-open & 37.19 & 95.83 & 0.517 & 0.047 \\

bottle-pour & 33.12 & 91.94 & 0.535 & 0.060
& bottle-wrap\_grasp & 7.09 & 60.32 & 0.703 & 0.144 \\

bowl-contain & 29.56 & 95.45 & 0.721 & 0.088
& bowl-pour & 12.37 & 90.57 & 0.817 & 0.103 \\

bowl-wrap\_grasp & 4.50 & 67.85 & 0.745 & 0.135
& chair-move & 15.86 & 80.17 & 0.559 & 0.123 \\

chair-sit & 46.93 & 96.48 & 0.751 & 0.063
& chair-support & 30.33 & 92.40 & 0.716 & 0.073 \\

clock-display & 13.74 & 86.73 & 0.513 & 0.071
& dishwasher-contain & 36.78 & 93.72 & 0.632 & 0.083 \\

dishwasher-open & 24.67 & 91.86 & 0.354 & 0.048
& display-display & 17.48 & 83.60 & 0.472 & 0.113 \\

door-open & 24.93 & 84.82 & 0.345 & 0.032
& door-pull & 27.50 & 92.49 & 0.318 & 0.029 \\

door-push & 12.30 & 82.82 & 0.555 & 0.066
& earphone-grasp & 4.05 & 53.95 & 0.199 & 0.197 \\

earphone-listen & 24.49 & 93.30 & 0.638 & 0.082
& faucet-grasp & 12.20 & 77.09 & 0.593 & 0.137 \\

faucet-open & 30.05 & 94.50 & 0.579 & 0.039
& hat-grasp & 3.93 & 42.74 & 0.508 & 0.157 \\

hat-wear & 8.39 & 67.53 & 0.582 & 0.131
& keyboard-press & 6.98 & 82.93 & 0.150 & 0.069 \\

knife-cut & 12.78 & 94.64 & 0.760 & 0.102
& knife-grasp & 21.54 & 87.26 & 0.690 & 0.081 \\

knife-stab & 20.91 & 99.54 & 0.756 & 0.021
& laptop-display & 2.48 & 39.88 & 0.067 & 0.171 \\

laptop-press & 21.16 & 94.42 & 0.559 & 0.050
& microwave-contain & 18.92 & 96.09 & 0.444 & 0.103 \\

microwave-open & 35.49 & 89.94 & 0.434 & 0.011
& mug-contain & 14.38 & 84.40 & 0.621 & 0.097 \\

mug-grasp & 42.05 & 93.93 & 0.499 & 0.068
& mug-pour & 5.16 & 82.91 & 0.620 & 0.082 \\

mug-wrap\_grasp & 1.13 & 42.83 & 0.357 & 0.163
& refrigerator-contain & 26.13 & 92.32 & 0.512 & 0.087 \\

refrigerator-open & 25.16 & 97.39 & 0.427 & 0.029
& scissors-cut & 20.10 & 96.98 & 0.445 & 0.120 \\

scissors-grasp & 2.59 & 51.34 & 0.392 & 0.154
& scissors-stab & 12.66 & 98.64 & 0.308 & 0.045 \\

storagefurniture-contain & 34.88 & 88.03 & 0.605 & 0.075
& storagefurniture-open & 17.89 & 91.65 & 0.332 & 0.030 \\

table-move & 5.12 & 60.25 & 0.563 & 0.181
& table-support & 15.86 & 85.66 & 0.723 & 0.094 \\

trashcan-contain & 24.29 & 82.64 & 0.377 & 0.133
& trashcan-open & 12.09 & 79.58 & 0.280 & 0.074 \\

trashcan-pour & 2.78 & 40.50 & 0.134 & 0.198
& vase-contain & 16.90 & 79.18 & 0.503 & 0.122 \\

vase-pour & 12.73 & 77.57 & 0.584 & 0.109
& vase-wrap\_grasp & 6.93 & 66.32 & 0.632 & 0.136 \\
\bottomrule
\end{tabular}
}
\end{table}

\paragraph{Failure Analysis.}
Table~\ref{tab:laso_unseen_per_class} indicates that wrap-grasping remains challenging, particularly for mugs, bowls, and bottles. One possible factor is ambiguity in the ground-truth annotations: multiple contact configurations can be physically plausible, while the annotated affordance map may capture only some of them. Consequently, a reasonable prediction may still differ from the reference in its location or coverage. Moreover, wrap-grasping involves an extended contact surface rather than a localized functional part, making it difficult to accurately match both the spatial extent and the point-wise score distribution. This is reflected in bowl-wrap-grasp and bottle-wrap-grasp, where relatively high SIM scores (0.745 and 0.703) coexist with low aIoU scores (4.50 and 7.09), suggesting partial distributional agreement without precise spatial overlap. Limited visibility across rendered views may further leave parts of these surrounding contact surfaces insufficiently supported by visual evidence. These cases suggest room for improving spatial coverage and handling ambiguity in interaction regions.

\end{document}